\documentclass{article}
\usepackage{xspace}
\usepackage{graphicx}
\usepackage{microtype}
\usepackage{subfigure}
\usepackage{booktabs} 
\usepackage{enumitem}
\usepackage{biblatex}
\usepackage{algorithm}
\usepackage{algorithmic}

\usepackage{amsmath}
\usepackage{amssymb}
\usepackage{mathtools}
\usepackage{amsthm}
\usepackage{hyperref}       
\usepackage[capitalize,noabbrev]{cleveref}

\usepackage[textsize=tiny]{todonotes}
\usepackage{multirow}

\usepackage{color,soul}

\newcommand{\hsk}[1]{{\color{black}#1}}

\usepackage{wrapfig}

\usepackage[preprint,nonatbib]{neurips_2022}

\usepackage[utf8]{inputenc} 
\usepackage[T1]{fontenc}    

\usepackage{url}            
\usepackage{booktabs}       
\usepackage{amsfonts}       
\usepackage{nicefrac}       
\usepackage{microtype}      
\usepackage{xcolor}         

\title{Federated Semi-Supervised Learning with Prototypical Networks}

\author{%
  Woojung Kim\\
  Graduate School of Data Science\\
  Seoul National University\\
  \texttt{kimwj94@snu.ac.kr} \\
  \And
  Keondo Park\\
  Graduate School of Data Science\\
  Seoul National University\\
  \texttt{gundo0102@snu.ac.kr} \\
  \And
  Kihyuk Sohn\\
  Google Research\\
  \texttt{kihyuks@google.com} \\  
  \And
  Raphael Shu\\
  AWS AI Labs\\
  \texttt{zhongzhu@amazon.com} \\
  \And
  Hyung-Sin Kim\\
  Graduate School of Data Science\\
  Seoul National University\\
  \texttt{hyungkim@snu.ac.kr} \\
}

\DeclareUnicodeCharacter{1EF3}{\`y}
\begin{document}

\maketitle

\newcommand{\proposal}{\texttt{ProtoFSSL}\xspace}

\begin{abstract}
\label{sec:abstract}
With the increasing computing power of edge devices, Federated Learning (FL) emerges to enable model training without privacy concerns. The majority of existing studies assume the data are fully labeled on the client side. In practice, however, the amount of labeled data is often limited. Recently, federated semi-supervised learning (FSSL) is explored as a way to effectively utilize unlabeled data during training.
%
In this work, we propose {\proposal}, a novel FSSL approach based on \textit{prototypical networks}. In \proposal, clients share knowledge with each other via \textit{lightweight prototypes}, which prevents the local models from diverging. For computing loss on unlabeled data, each client creates accurate pseudo-labels based on shared prototypes. Jointly with labeled data, the pseudo-labels provide training signals for local prototypes. Compared to a FSSL approach based on weight sharing, the prototype-based inter-client knowledge sharing significantly reduces both communication and computation costs, enabling more frequent knowledge sharing between more clients for better accuracy.
%
%
%
%
%
In multiple datasets, \proposal results in higher accuracy compared to the recent FSSL methods with and without knowledge sharing, such as  FixMatch, FedRGD, and FedMatch. On SVHN dataset, \proposal performs comparably to fully supervised FL methods\footnote{The code is available at https://github.com/kimwj94/ProtoFSSL.}. 

\end{abstract}

\vspace{-1ex}
\section{Introduction}
\label{sec:introduction}
\vspace{-1ex}

Federated Learning (FL) \cite{mcmahan2017communication} is a machine learning framework that has been actively studied recently. 
FL leverages the computing power of clients to enable local model training without data sharing, helping to resolve privacy concerns.
Mainstream approaches, such as FedAvg \cite{mcmahan2017communication} or FedProx \cite{li2020federated}, ask clients to send locally trained models to a server, while the server updates a global model using a certain aggregation method. Once the global model is updated, the global weights are broadcasted back to the clients, completing one round of training. The rounds repeat until reaching convergence.

Most of the existing FL methods focus on supervised learning (SL) setting, where the client-side data are fully labeled \cite{he2020group, zhu2021data, li2021model}. However, given that labeled data are costly to obtain, it is practical to assume that the client-side data are mostly unlabeled. 
%
\emph{Federated Semi-Supervised Learning} (FSSL) is proposed to harvest value from the unlabeled client-side data.
While a number of methods for semi-supervised learning (SSL), such as FixMatch \cite{sohn2020fixmatch} and UDA \cite{NEURIPS2020_44feb009}, have been proposed, a naïve application to FSSL that simply replaces the local model training with SSL methods is shown less successful. This is because clients have limited amount of data and heterogeneous (non-i.i.d.) data distribution, causing high gradient diversity among local models~\cite{zhang2020improving}.

FedMatch \cite{jeong2021federated} was proposed to address the aforementioned issue by 
enabling a client to exploit knowledge from other clients when using unlabeled data, hence preventing local model divergence. 
However, its method for inter-client knowledge sharing in the form of \textit{model weights} increases communication and computation overhead on clients, linearly proportional to the model size. Given that clients are much more resource-constrained than the server, increasing client-side burden is not desirable in FSSL.
\hsk{Furthermore, recent  studies~\cite{zhang2020improving,lin2021semifed} show that it provides significantly lower accuracy than other FSSL methods that do not share knowledge between clients, suggesting that the idea of knowledge sharing may be ineffective; to share or not is an interesting question to investigate.}

\begin{figure}[t]
    \centering
    \includegraphics[width=\linewidth]{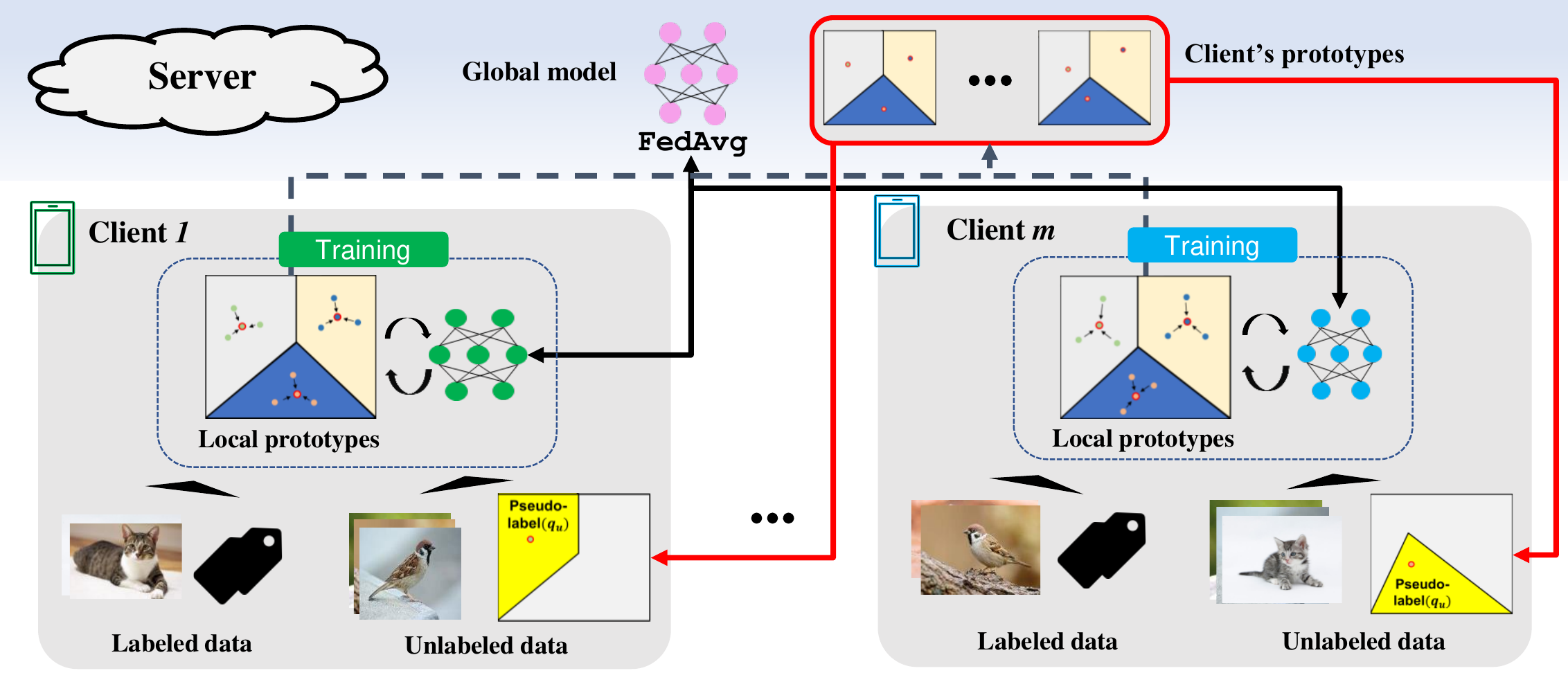}
    \vspace{-2ex}
    \caption{Overall design of \proposal. We used \emph{prototypes} to train the network, and share inter-client knowledge to produce \emph{psuedo-labels} for unlabeled data.}
    \vspace{-2ex}
    \label{fig:overallDesign}
\end{figure}

In this work, we revisit inter-client knowledge sharing and develop a tailored approach that shares knowledge between clients \textit{without using model weights} for effective inter-client consistency regularization. Specifically, we propose \proposal that adopts prototypical network~\cite{snell2017prototypical}, originally developed for metric-based meta learning, to FSSL. In \proposal, inter-client knowledge is shared in the form of \textit{prototypes} that are much lighter than model weights. Each client utilizes the shared prototypes for \textit{pseudo-labeling} its unlabeled data with \textit{consistency regularization}, which improves pseudo-labeling quality and mitigates local model divergence in non-i.i.d. data environments. Given the light communication cost of prototypes, \proposal allows more clients to share knowledge more frequently for better model convergence. The overall design of our system is illustrated in Figure~\ref{fig:overallDesign}.

The contributions of this work are as follows:
\vspace{-1.5ex}
\begin{itemize}[leftmargin=*]
  \item We investigate FSSL, a practical application of FL with partially labeled data on the clients. Specifically, we propose \proposal, a novel method that \hsk{combines prototypes, inter-client knowledge sharing, and consistency regularization together with FL for efficient and accurate model training.}
  

  \item We design two lightweight components to utilize unlabeled data: (1) prototype-sharing and (2) inter-client prototype-based pseudo-labeling. \hsk{Even in lightweight ResNet9}, these components achieve 2x less communication than FedMatch and \hsk{1.7$\sim$3.4x less client-side computation than FedMatch and other data augmentation-based consistency regularization methods.} Furthermore, they lead to accuracy improvement with more frequent knowledge sharing among more clients. 
 
  \item Experiments show that on CIFAR-10, SVHN, and STL-10 datasets, \proposal outperforms recent FSSL algorithms, both with and without knowledge sharing, by 2.9\%$\sim$5.7\% in accuracy. On SVHN, \proposal is even comparable to fully supervised learning. \hsk{The results verify that inter-client knowledge sharing is still effective, if accompanied with a carefully designed mechanism.}

  \item We show that \proposal is complementary to other approaches for FSSL that prevent local gradient diversity  without using inter-client knowledge, such as various normalization methods.  
\end{itemize}
\vspace{-1.5ex}
\vspace{-1ex}
\section{Related work}
\vspace{-1.5ex}
\textbf{Federated Learning (FL).} 
FL was first introduced in~\cite{mcmahan2017communication} as a privacy-preserving decentralized training that leverages local clients' data as well as computation power. Communication and computation efficiency, and data heterogeneity have been pointed as the major challenges in FL~\cite{li2020federatedSurvey}. 
A number of studies have tried to reduce communication burden, such as model compression~\cite{konevcny2016federated, alistarh2017qsgd, ivkin2019communication}, asynchronous update \cite{chen2019communication}, and  matched averaging~\cite{wang2020federated}.
%
%
Data heterogeneity across clients deteriorate the convergence of FL, which has been investigated in various methods, such as different batch normalization~\cite{li2021fedbn, diao2021heterofl}, regularization for local optimization \cite{li2020federated, acar2021federated}, meta learning or multi-task learning~\cite{NEURIPS2019_f4aa0dd9}, and knowledge distillation~\cite{li2019fedmd}.

Similarly, this work focuses on data heterogeneity and communication/computation burden on clients. While inter-client prototype sharing is proposed as a lightweight method to improve convergence, its interaction with various normalization methods is also evaluated, showing synergy between the two.

\textbf{Semi-supervised learning (SSL).} 
%
Given that labeling all data is very costly, time-consuming and sometimes requires domain expertise, researchers have tried to develop learning techniques which tackle label deficiency, such as consistency regularization and pseudo-labeling. 

As the name implies, consistency regularization~\cite{sajjadi2016regularization} imposes model predictions from input perturbations or stochastic output~\cite{NIPS2015_378a063b, bachman2014learning} to be consistent. 
Many approaches were suggested recently, including 
data augmentation in UDA~\cite{NEURIPS2020_44feb009} and ReMixMatch~\cite{berthelot2019remixmatch} to enforce consistency between weakly and strongly augmented unlabeled examples. 
Pseudo-labeling produces either soft~\cite{DBLP:conf/iclr/LaineA17,NIPS2017_68053af2,NEURIPS2019_1cd138d0} or hard~\cite{sohn2020fixmatch} labels for unlabeled data using a model prediction. To achieve accurate pseudo-labeling, 
MixMatch~\cite{NEURIPS2019_1cd138d0} sharpens pseudo-labels by producing low-entropy labels for data-augmented unlabeled examples and mixing labeled and unlabeled data. FixMatch~\cite{sohn2020fixmatch} pseudo-labels an unlabeled sample only when the model predicts with high confidence and applies consistency regularization. \hsk{However, applying these augmentation-based pseudo-labeling in FL causes more computation on clients.}

\textbf{Federated Semi-Supervised Learning (FSSL).}
Given that labeling data at local clients is even harder than at the server~\cite{jin2020towards}, SSL and FL scenarios fit well together, leading to a relatively new problem called Federated Semi-Supervised Learning (FSSL). Two FSSL scenarios were introduced in~\cite{jeong2021federated}: (1) \textit{labels-at-client} where clients have limited labels while the server has no data and (2) \textit{labels-at-server} where client data is completely unlabeled while the server has labeled data.
FedMatch~\cite{jeong2021federated}, pioneering work in this regime, proposes a disjoint learning scheme in which the model is divided into two sets of weights for supervised learning and unsupervised learning, respectively. For unsupervised learning, knowledge is shared between clients in the form of \textit{model weights} and inter-client consistency loss is used for consistency regularization. However, training two sets of parameters separately and sharing weights between clients increases computation and communication overhead on clients.

Some recent work focuses on the labels-at-server scenario. \hsk{SSFL~\cite{he2021ssfl} tackles client-side label deficiency by using SimSiam~\cite{chen2021exploring} for local training.} FedRGD~\cite{zhang2020improving} showed that gradient diversity of consistency regularization loss is reduced with group normalization as well as grouping-based model averaging. SemiFL~\cite{lin2021semifed} alternates training between the server and the clients and utilizes static batch normalization to enhance the model.
Importantly, FedRGD and SemiFL reported that FedMatch significantly underperforms FSSL methods that do not utilize inter-client knowledge, implying that knowledge sharing \hsk{may not be helpful for model training.} 

In contrast, this work focuses on the labels-at-client scenario, \hsk{investigates knowledge sharing process thoroughly, and presents a new effective approach} to enforce inter-client consistency regularization in the presence of data heterogeneity, in a more accurate and communication/computation efficient way.

\textbf{Prototypical Networks.}
Prototypical network~\cite{snell2017prototypical} trains a model that produces a reasonable prototype (\hsk{low-dimensional} embedding vector) for each class and classifies a new data by using distance between its embedding vector and each prototype. The concept of prototype has been successfully applied in various areas, such as meta learning~\cite{snell2017prototypical,allen2019infinite, kim2019variational}, domain adaptation~\cite{pan2019transferrable, toldo2021unsupervised}, and semantic segmentation~\cite{wang2019panet, dong2018few}. 
The authors in \cite{ren2018meta} extend prototypical network for SSL scenarios by incorporating unlabeled data to produce prototypes. There have been recent attempts to apply prototypes to resolve data heterogeneity among clients in FL settings~\cite{tan2021fedproto, michieli2021prototype}. 
As another step forward, given that prototypes fit well when labeled samples are limited~\cite{snell2017prototypical}, we utilize the prototypes to represent client knowledge in FSSL scenarios. To the best of our knowledge, our proposal \proposal is the first tailored application of prototypes in FSSL.
\vspace{-2ex}
\section{Backgrounds}
\vspace{-1.5ex}

This section introduces three important backgrounds that \proposal builds upon: federated learning, prototypical networks, and pseudo-labeling, which will be merged into the FSSL problem in Section~\ref{sec:proposal}.

\vspace{-1.5ex}
\subsection{Supervised Federated Learning}
\vspace{-1ex}

\texttt{FedAvg}~\cite{mcmahan2017communication}, a representative FL method operates as follows. Each training round $r$ consists of broadcasting, local training, and aggregation. In broadcasting step, the server selects a set of $m$ clients, $M_r$, and sends the global model $\theta^{r-1}$ to the clients in $M_r$. Each client $i (\in M_r)$ locally updates $\theta^{r-1}$ to $\theta^{r}_i$ using its own data and ground-truth labels $D_i$ for multiple epochs $E$. The clients send the local models $\theta^r_i$ back to the server and the server updates the global model as
\vspace{-0.5ex}
\begin{equation}
    \label{eqn:fedAvg}
    \theta^{r} = \sum\limits_{i \in M_r} \frac{|D_i|}{\sum_{i \in M_r}{|D_i|}} {\theta^r_i}
\end{equation}
Importantly, client datasets are assumed to have heterogeneous (non-i.i.d.) distribution.

\subsection{Supervised Prototypical Networks}
\label{sec:protonet}
\vspace{-1ex}

Prototypical network \cite{snell2017prototypical} is trained to provide a good embedding space and a prototype for each class in the embedding space. The network converts a data sample into an embedding vector and classifies the vector using distance from each prototype. %
Define \( K\) as the set of classes and $D_k$ as training dataset of class $k$ ($\in K$). The training set $D_k$ is divided into two parts: (1) \emph{support set} \(S_k\) that is a random subset of \(D_k\) and (2)  the remaining \textit{query set} \(Q_k\) (\( D_k\setminus S_k \)). Then prototype \(c_k\) for class $k$ is computed from support set as
\begin{equation}
    \label{eqn:prototype}
    c_k=\frac{1}{\left|S_k\right|}\sum_{x \in S_k}{f_\theta(x)}
\end{equation}
where \(f_\theta\) is a neural network parameterized by weights \(\theta\). Then, the model is trained with the loss: 
\begin{align}
    \label{eqn:prototypeLoss}
    \mathcal{L} = -\frac{1}{\left|K\right|\left|Q_k\right|}\sum_{k \in K}\sum_{x \in Q_k} \log \frac{\exp (-d (f_\theta (x), c_k))}{\sum\limits_{k^{\prime} \in K} \exp (-d (f_\theta (x), c_{k^{\prime}}))}
\end{align}
where \(d\) is Euclidean distance function. The model is trained so that embedding vectors of the same class are located close together, and those of different classes are located far away. \proposal combines the idea of prototype with federated semi-supervised learning.

\subsection{Pseudo-labeling for Unlabeled Data}
\vspace{-1ex}

Pseudo-labeling unlabeled data with consistency regularization is important for effective semi-supervised learning. 
For example in MixMatch, several augmented data $u_a$($a=1,2,...,A$) for one unlabeled data $u$ are used to make a robust pseudo-label. To this end, average of the model's prediction across $A$ augmentations of $u$, $p(u)$, is computed as
\begin{equation}
    \label{eqn:pseudoLabelMixMatch}
    p(u)=\frac{1}{A}\sum_{a=1}^{A}{f_\theta({u}_{a})}
\vspace{-1ex}
\end{equation}
Further, MixMatch \textit{sharpens} the distribution $p(u)$ to lower the entropy of guessed label as follows:
\begin{equation}
    \label{eqn:sharpening}
    \bar{p}_k(u)=\frac{p_{k}(u)^{\frac{1}{T}}}{\sum_{k' \in K}p_{k'}(u)^{\frac{1}{T}}}          
\end{equation}
where \(\bar{p}_{k}(u)\) is the probability for class $k$ and \(T\) is a hyperparameter called \textit{temperature}. The sharpened probability distribution $\bar{p}(u)$ is the \textit{soft} pseudo-label for data $u$. 
\hsk{Consistency regularization of \proposal is inspired by MixMatch but different in that it utilizes other clients' prototypes and requires \textit{zero data augmentation} on resource-constrained clients.}
\section{Prototypical Networks for FSSL}
\label{sec:proposal}

In this section, we describe details of \proposal that aims to provide efficient inter-client consistency regularization for FSSL by using lightweight prototypes as a form of inter-client knowledge. \cref{alg:ProtoFSSL} summarizes the training process and Figure~\ref{fig:overallDesign} illustrates \proposal architecture.

\vspace{-0.5ex}
\subsection{Overview}

We consider a FSSL problem where the server has no data but each client $i$ has a private non-i.i.d. dataset $D_i=D_i^L\cup{D_i^U}$ where \(D_i^L = \cup{\{D_{i,k}^L\}}_{k \in K}\) is the labeled dataset and $D_i^U$ is the unlabeled dataset. $K$ is defined as the set of class labels. For all clients, the unlabeled dataset is assumed to be much larger than the labeled dataset, i.e., \(\lvert{D_i^L}\rvert{\ll}\lvert{D_i^U}\rvert\).

With the data setting, the goal of this work is to train a prototypical network $f_\theta$, parameterized by $\theta$, in a federated manner. At each communication round $r$, for $E$ local epochs, each participating client $i$($\in M_r$) locally updates global parameters $\theta^{r-1}$ to $\theta^r_i$ using its labeled and unlabeled datasets, and also updates its local prototype $c_{i,k}$ for each class $k$($\in K$). 
In each local epoch, client $i$ randomly samples a labeled support set $S^L_{i,k}$ from $D^L_{i,k}$ and a labeled query set $Q^L_{i,k}$ from $D^L_{i,k} \setminus{S^L_{i,k}}$ for each class $k$, and also randomly samples an unlabeled query set $Q^U_i$ from $D^U_i$.

Note that \proposal utilizes the client-specific prototypes to enable knowledge sharing between clients. Specifically, at each round $r$, the server selects a set of helper clients $H_r$ and shares the helpers' local prototypes $c_{j,k}$($j \in H_r$) with each client $i$ in $M_r$.

\begin{algorithm}[t]
   \caption{ProtoFSSL}
   \label{alg:ProtoFSSL}
\begin{algorithmic}
    \STATE {\bfseries RunServer():}
    \STATE Initialize global parameter $\theta^0$
    \FOR{each round $r = 1, 2, ..., R$}
    \STATE $M_r \leftarrow$ random sample of $m$ clients
    \STATE Select helper set $H_r$ from previous active clients $M_{r-1}$
    \FOR{each client $i \in M_r$ {\bfseries in parallel}}
    \STATE $\theta^r_i, \{c_{i,k}\}_{k \in K} \leftarrow  $RunClient$(\theta^{r-1}, \{c_{j,k}\}_{k \in K, j \in H_r})$
    \ENDFOR
    \STATE Update global model $\theta^r$ using $\{\theta^r_i\}_{i \in M_r}$ as in Eq.\eqref{eqn:fedAvg}
    \STATE Store prototypes $\{c_{i,k}\}_{k \in K, i \in M_r}$
    \ENDFOR
    \STATE
    \STATE {\bfseries RunClient}($\theta^{r-1}, \{c_{j,k}\}_{k \in K, j \in H_r}$):
    \STATE Initialize $\theta^r_{i,0} \leftarrow \theta^{r-1}$
    \FOR{each local epoch $e = 1, 2, ..., E$}
    \FOR{each class $k \in K$}
        \STATE $ S^L_{i,k} \leftarrow$ random sample from $D^L_{i,k}$ 
        \STATE $ Q^L_{i,k} \leftarrow$ random sample from $D^L_{i,k}\setminus{S^L_{i,k}}$
        \STATE $ c_{i,k} \leftarrow $MakePrototype$(\theta^r_{i,e-1}, S^L_{i,k})$ 
    \ENDFOR

    \STATE $Q^U_i \leftarrow$ random sample from $D^U_i$
    \STATE $\{\bar{p}(u)\}_{u\in{Q^U_i}} \leftarrow $\textbf{PseudoLabel}$(Q^U_i, \{c_{j,k}\}_{k \in K, j \in H_r})$
    \STATE $\theta^r_{i,e} \leftarrow \theta^r_{i,e-1} - \eta\nabla_{\theta}loss_i$ (loss function in Eq.\eqref{eqn:loss})
    \ENDFOR
    \STATE $\{c_{i,k}\}_{k \in K} \leftarrow $MakePrototype$(\theta^r_{i,E}, \{D^L_{i,k}\}_{k \in K})$
    \STATE Send $\theta^r_{i,E}$, $\{c_{i,k}\}_{k \in K}$ back to server 
\end{algorithmic}
\end{algorithm}

\subsection{Pseudo-Labeling with Inter-Client Knowledge}

\begin{figure}[t]
    \centering
    \subfigure[Pseudo-labeling with external prototypes from helper clients]{
    \centering
        \includegraphics[width=.8\linewidth]{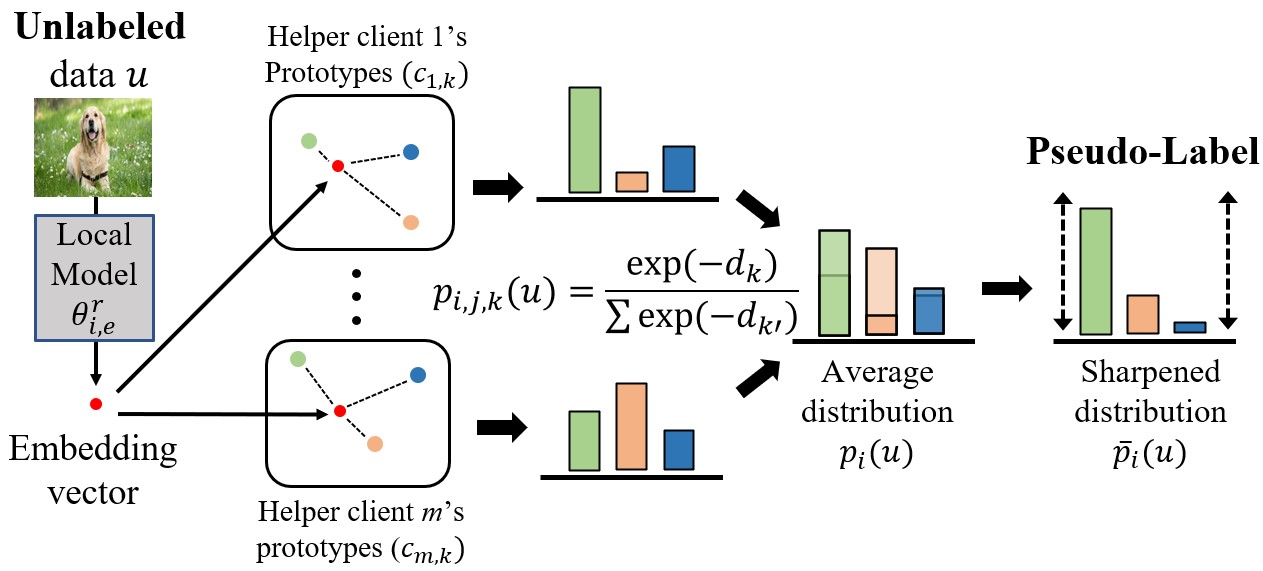}
        \label{fig:pseudolabel}
    }
    \subfigure[Loss calculation using both labeled and unlabeled datasets with internal prototypes]{
    \centering
        \includegraphics[width=.8\linewidth]{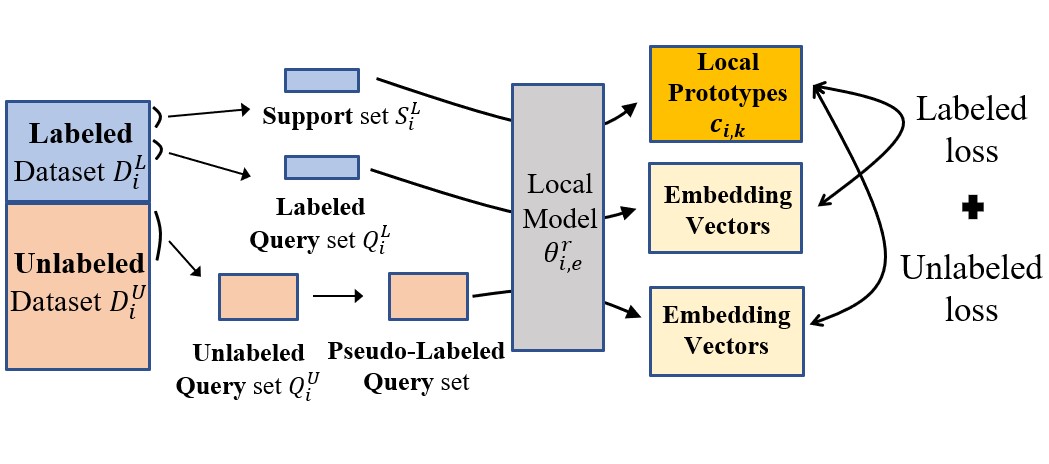}
        \label{fig:loss}
    }
    \vspace{-1ex}
    \caption{Key design components of \proposal for local training on each client $i$.}
    \vspace{-2ex}
    \label{fig:localtraining}
\end{figure}

As depicted in Figure~\ref{fig:pseudolabel}, for local training on unlabeled data in each epoch, client $i$ in $M_r$ pseudo-labels its unlabeled query data $u$($\in Q^U_i$) using external prototypes from helper clients in $H_r$.
Assuming that $\theta^r_{i,e}$ is the current weights that client $i$ has in local epoch $e$, client $i$ first computes an embedding vector $f_\theta(u)$ for an unlabeled datapoint $u$ using the local weights $\theta^r_{i,e}$. Then Euclidean distance is computed between the embedding vector $f_\theta(u)$ and prototypes \(c_{j,k}\) for each class $k$($\in K$) and helper client $j$($\in H_r$). With respect to helper client $j$'s prototypes, the probability for unlabeled data $u$ to be class $k$, denoted as $p_{i,j,k}(u)$, is computed using softmax of the negative of the distance for each class as
\begin{align}
    \label{eqn:other_client_probability}
    p_{i,j,k}(u) = \frac{\exp\{-d(f_\theta(u),\ c_{j,k})\}}{\sum_{k' \in K}\exp\{-d(f_\theta(u),\ c_{j,k'})\}}
\end{align}

After computing class probabilities for every helper client's prototypes, average class probability for $u$ is computed as 
\begin{equation}
    \label{eqn:pseudo-labeling}
    p_{i,k}(u) = \frac{1}{|H_r|}\sum_{j \in H_r} p_{i,j,k}(u)
\end{equation}
Sharpening $p_{i,k}(u)$ to $\bar{p}_{i,k}(u)$ as in Eq.\eqref{eqn:sharpening} results in the final class probability distribution $\bar{p}_i(u)$, which is used for a soft pseudo-label for data $u$.

While inspired by MixMatch, the main difference in our pseudo-labeling technique is that probability distributions are averaged over helper clients, instead of data augmentations, which is appropriate for resolving heterogeneous data distributions over clients in FSSL scenarios. 
Furthermore, while MixMatch needs to run the model $A$  times (the number of augmentations) for pseudo-labeling a single data, \proposal does it only once, reducing computation burden of resource-constrained clients.  

\vspace{-1ex}
\subsection{Loss}
\vspace{-1ex}
After pseudo-labeling each datapoint in $Q^U_i$, client $i$ calculates the loss using both labeled and pseudo-labeled data, as depicted in Figure~\ref{fig:loss}. 
For each epoch $e$, client $i$ produces its local prototypes for each class $k$ using its labeled support set $S^L_{i,k}$($\subset{D^L_{i,k}}$) and the current weights $\theta^r_{i,e}$, as in Eq.\eqref{eqn:prototype}. Then the model predictions for each labeled data $x$($\in Q^L_{i,k}$) and unlabeled data $u$($\in Q^U_{i}$) are given as in Eq.\eqref{eqn:other_client_probability}, resulting in class probabilities $p_{i,i,k}(x)$ and $p_{i,i,k}(u)$ for each class $k$, respectively. Aggregation of these probabilities for all classes produces probability distributions $p_{i,i}(x)$ and $p_{i,i}(u)$. Finally, the loss in each local epoch is computed using cross entropy loss as follows:
\vspace{-1ex}
\begin{align}
    \label{eqn:loss}
    \mathcal{L}_i &= \frac{1}{\sum_{k\in K}{|Q_{i,k}^{L}|}} \sum_{k \in K}\sum_{x\in{Q^L_{i,k}}}{\mathrm{CroE.}(y_k,\ p_{i,i}(x))} + \lambda \cdot \frac{1}{|Q^U_i|}  \sum_{u\in{Q^U_i}}{\mathrm{CroE.}(\bar{p}_i(u),\ p_{i,i}(u))}
\end{align}
where \(\lambda\) is a hyperparameter that regularizes the weight of unlabeled loss and $y_k$ is a ground-truth label for class $k$.

\vspace{-1ex}
\subsection{Prototype Sharing}
\vspace{-1ex}

After finishing local training for $E$ epochs, each client computes new local prototypes as local knowledge to share with others. To this end, a client produces embedding vectors using the updated local weights $\theta^r_i$, from all the labeled dataset $D^L_{i,k}$ for each class $k$ (i.e., not a random sample). Each client sends both the local prototypes and local model weights to the server, becoming a potential helper client in the next round $r+1$.

\vspace{-1ex}

\begin{table}[t]
    \centering
    \caption{Computation and communication costs of a client in each round: Analysis.}
    \label{table:analysis}
    \resizebox{\linewidth}{!}{%
    \begin{tabular}{lccc}
    \toprule
    Methods & Comp. cost & Comm.cost\\
    \midrule
    MixMatch-FedAvg & \(F(|D^L_i|+A|D^U_i|)E\)     & \(2\Theta\)           \\
    FedRGD (or FixMatch-FedAvg)         & \(F(|D^L_i|+2|D^U_i|)E\)     & \(2\Theta\)           \\
    FedMatch        & \(F(|D^L_i|+(2+|H_r|)|D^U_i|)E\) & \((4+2|H_r|/R_H)\Theta\)    \\
    \multirow{2}{*}{\proposal}       & $F(|D^L_i| + |D^U_i|)E+$   & $2\Theta+$           \\
           & $\textbf{C}|H_r||K||D^U_i|E + F|D^L_i|$   & $\textbf{C}(1+|H_r|)|K|$           \\
    \bottomrule
    \end{tabular}}
\end{table}

\begin{table}[t]
    \caption{Per-client computation and communication costs in each round: An example when using ResNet9 and CIFAR-10 as in Table~\ref{table:accuracies}.}
    \label{table:analysis_ex}
    \centering
    \resizebox{.8\linewidth}{!}{%
    \begin{tabular}{lccc}
    \toprule
    Methods & Comp. cost & Comm.cost\\
    \midrule
    MixMatch-FedAvg & 782.0 GFLOP     & 52.6 MB           \\
    FedRGD (or FixMatch-FedAvg) & 782.0 GFLOP     & 52.6 MB           \\
    FedMatch        & 1526.0 GFLOP & 115.6 MB    \\
    \proposal       & 447.9 GFLOP & 52.6 MB           \\
    
    \bottomrule
    \end{tabular}}
\end{table}

\section{Cost Analysis}

We analyze computation and communication overhead of \proposal, which is important for resource-constrained clients.
While maintaining the notations used in previous sections, we assume that $\Theta$ is the size of model $\theta$, $F$ is computation cost to run the model $\theta$ for a single datapoint, $C$ is the size of a prototype, and $R_H$ is round interval to update helper clients (for FedMatch). For fair comparison, \proposal is assumed to consume the entire labeled and unlabeled datasets in an epoch, instead of using random-sampled subsets. 
Table~\ref{table:analysis} shows the results where MixMatch-FedAvg and FixMatch-FedAvg run the SSL methods MixMatch and FixMatch directly on a client. Note that a recent FSSL method FedRGD~\cite{zhang2020improving} also uses FixMatch for pseudo-labeling on clients.

To analyze the results, note that the costs highly depend on model-related parameters $F$ and $\Theta$($\gg C$) and the amount of unlabeled data $|D^U_i|$($\gg|D^L_i|$). 
Regarding computation cost, the costs of MixMatch-FedAvg and FedMatch significantly increase with the number of augmentations $A$ and helper clients $|H_r|$, respectively, which makes these methods hard to scale. Without these varying parameters, FedRGD (or FixMatch-FedAvg) provides consistent computation overhead. On the other hand, \proposal provides the lowest computation cost since the two terms $C|H_r||K||D^U_i|E$ and $F|D^L_i|$ are much smaller than others. Although $C|H_r||K||D^U_i|E$ contains unlabeled dataset size $|D^U_i|$, the term is still negligible as simple distance calculation related to $C$ rather than a model computation $F$.

As for communication costs, all the methods require at least $2\Theta$ to exchange global and local models between the server and a client. FedMatch requires additional cost to share model weights between clients, which increases proportional to $|H_r|$ and limits scalability of the method.  In contrast, \proposal's additional cost for knowledge sharing is small since it depends on the prototype size $C$ instead of model size $\Theta$.

As an example, consider using ResNet9 in CIFAR-10 where unlabeled data is 9.8 times more than labeled data, as in experiments in~\cite{zhang2020improving,jeong2021federated}. In this setting, $C=2 kB$ is 13,000 times smaller than $\Theta = 26 MB$. Table~\ref{table:analysis_ex} shows the results, assuming that $|H_r|=2$ as in~\cite{jeong2021federated} and $A=2$ as in~\cite{NEURIPS2019_1cd138d0}. \proposal reduces computation 71\% compared to FedMatch and even 43\% compared to FixMatch that is known as a lightweight SSL method. \proposal's communication burden is less than half of FedMatch and similar to MixMatch and FedMatch.
Overall, \proposal enables inter-client knowledge sharing and pseudo-labeling in a significantly lighter way than the existing methods.
\vspace{-1ex}
\begin{table*}
    \centering
    \caption{Comparison of test accuracy to other methods from different datasets.}
    \label{table:accuracies}
    \resizebox{\textwidth}{!}{%
    \begin{tabular}{c|ccc|ccc|cc}
    \hline
                        & \multicolumn{3}{c|}{CIFAR-10} & \multicolumn{3}{c|}{SVHN} & \multicolumn{2}{c}{STL-10}     \\
    \hline
    \multirow{2}{*}{Method}     & \multirow{2}{*}{L:U}      & \multicolumn{2}{c|}{Test Acc. (\%)} 
    & \multirow{2}{*}{L:U}      & \multicolumn{2}{c|}{Test Acc. (\%)} 
    & \multirow{2}{*}{L:U}      & \multirow{2}{*}{Test Acc. (\%)}        \\
    
                                &                           & IID & Non-IID
                                &                           & IID & Non-IID
                                &                           &                                       \\
    \hline
    FedAvg-SL                   & \multirow{2}{*}{540:0}    & 81.7  & 80.0
                                & \multirow{2}{*}{540:0}    & 93.4  & 91.7
                                & \multirow{2}{*}{-}        & \multirow{2}{*}{-}                    \\
    
    FedProx-SL                  &                           & 83.0  & 81.2
                                &                           & 95.3  & 93.9
                                &                           & \multirow{2}{*}{-}                    \\
    \hline
    FedAvg                      & \multirow{2}{*}{50:0}     & \multicolumn{2}{c|}{62.2}
                                & \multirow{2}{*}{50:0}     & \multicolumn{2}{c|}{83.5}
                                & \multirow{2}{*}{100:0}    & 72.3                                  \\
    
    FedProx                     &                           & \multicolumn{2}{c|}{62.7}
                                &                           & \multicolumn{2}{c|}{83.9}
                                &                           & 72.0                                  \\
    \hline
    FixMatch-FedAvg             & \multirow{5}{*}{50:490}   & 62.4  & 61.8
                                & \multirow{5}{*}{50:490}   & 86.5  & 87.0
                                & \multirow{5}{*}{100:980}  & 73.7                                  \\
    
    FixMatch-FedProx            &                           & 63.4  & 61.6
                                &                           & 87.6  & 87.5
                                &                           & 73.6                                  \\
    
    FedMatch~\cite{jeong2021federated}                    &                           & 57.7  & 59.1
                                &                           & 82.9  & 81.9
                                &                           & 66.1                                  \\
    
    \proposal - FedAvg          &                           & 67.8  & 66.5
                                &                           & 88.3  & 88.1
                                &                           & 76.4                                  \\
    
    \proposal - FedProx         &                           & \textbf{68.1}  & \textbf{67.6}
                                &                           & \textbf{88.6}  & \textbf{88.8}
                                &                           & \textbf{76.6}                                  \\
                                
    \hline
    FedRGD~\cite{zhang2020improving}                      &  \multirow{3}{*}{50:490}  & 69.7  & 63.4
                                &  \multirow{3}{*}{50:490}  & -  & -
                                &  \multirow{3}{*}{100:980} & -                                  \\
    
    \proposal(with BN) - FedAvg &                           & \textbf{73.9}  & \textbf{73.4}
                                &                           & \textbf{93.8}  & \textbf{93.2}
                                &                           & 78.2                                  \\
                                
    \proposal(with BN) - FedProx &                           & \textbf{73.9}  & 72.5
                                &                           & 91.1  & 90.8
                                &                           & \textbf{79.3}                                  \\

    \hline
    \end{tabular}%
    }
\end{table*}

\section{Experiments}
\label{sec:experiment}
\vspace{-1ex}

We incorporate following baselines in experiments:
\vspace{-1ex}
\begin{itemize}[leftmargin=*]
  \item \textbf{FedAvg-SL} and \textbf{FedProx-SL}: Mainstream FL frameworks FedAvg and FedProx applied in a SL scenario where each client has fully labeled dataset. These methods provide upper-bound accuracy.
  
  \item \textbf{FedAvg} and \textbf{FedProx}: FedAvg and FedProx applied in a restricted SL scenario where each client has only a small labeled dataset $D^L_i$($\ll |D_i|$) without any unlabeled dataset $D^U_i$. These methods provide lower-bound accuracy.
  
  \item \textbf{FixMatch-FedAvg} and \textbf{FixMatch-FedProx}: Naïve combination of the state-of-the-art SSL method, FixMatch~\cite{sohn2020fixmatch}, with the two FL frameworks. Both labeled data \(D^L_i\) and unlabeled data \(D^U_i\) are utilized in clients. 
  
  \item \textbf{FedMatch} and \textbf{FedRGD}: Recently proposed FSSL methods. FedMatch tries to mitigate local model divergence using inter-client knowledge while FedRGD does it using grouped normalization and grouped-aggregation without knowledge sharing. Each client utilizes both \(D^L_i\) and \(D^U_i\). 
\end{itemize}

For fair comparison, we follow the setting in prior works on FSSL~\cite{jeong2021federated, zhang2020improving}. There is a total of 100 clients where 5 active clients ($m=5$) are randomly selected to participate in training in each round. ResNet9 is used for baselines but ResNet8 is used for \proposal since our method does not require the last fully connected layer.

We evaluate on three datasets: CIFAR-10~\cite{cifar10}, SVHN~\cite{svhn} and STL-10~\cite{stl10}. We split CIFAR-10 data into 54,000 training, 3,000 validation, and 3,000 test sets and SVHN data into 54,000 training, 2,000 validation, and 2,000 test sets. The training datapoints are distributed to 100 clients (i.e., 540 data per client) where 5 instances per class are labeled and the remaining 490 instances are unlabeled (i.e. \(|D_{i,k}^{L}|=5\), $|D_i^L|=50$ and \(|D^U_i|=490\)). For unlabeled data, we use an i.i.d setting where each client has same number of data per class and a non-i.i.d setting where each client has unbalanced class distribution as in~\cite{jeong2021federated}. 
For STL-10, each client has 1,080 datapoints where 100 datapoints are labeled and the other 980 are unlabeled. Since STL-10, originally constructed for SSL, does not provide class information for unlabeled data, we randomly distribute its unlabeled data to each client. Since STL-10 is not fully labeled, FedAVG-SL and FedProx-SL cannot be evaluated in the dataset.

\subsection{Experimental Results}
\vspace{-1ex}

\begin{figure}[t]
    \centering
    \includegraphics[width=.9\linewidth]{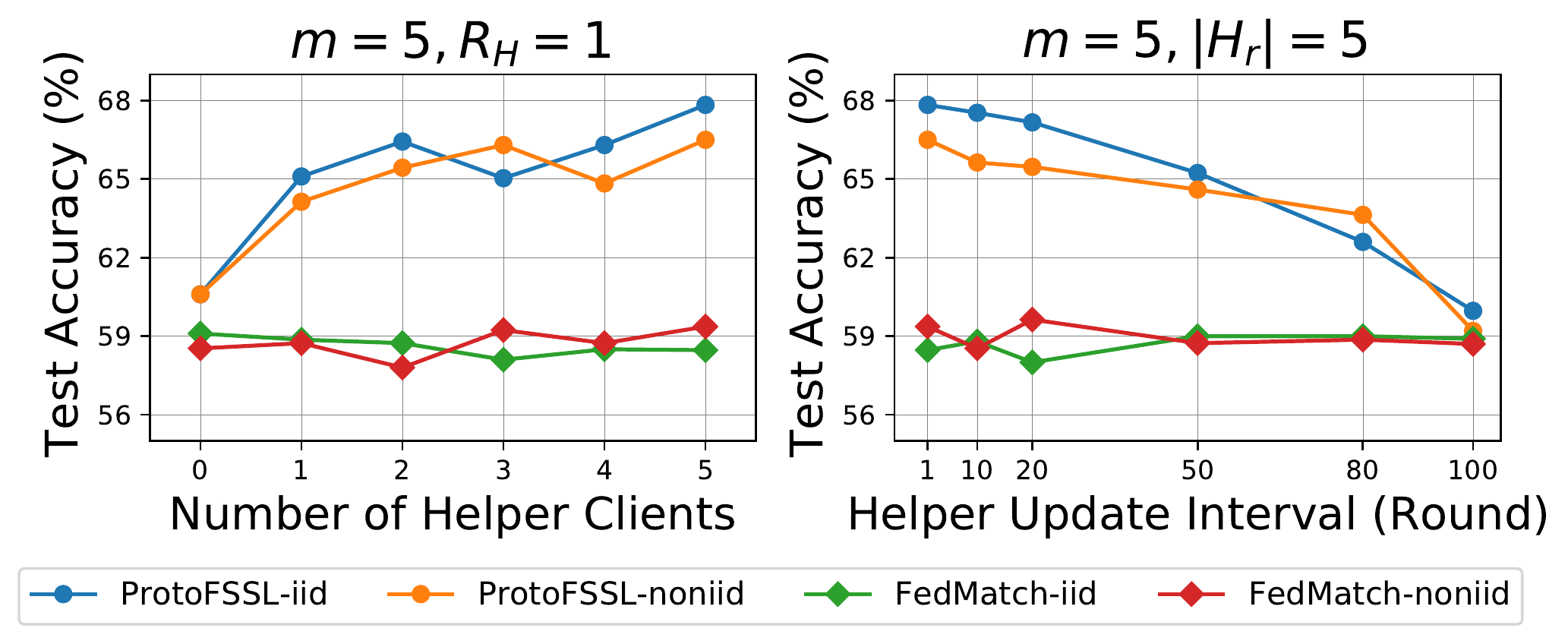}
    \caption{Test Accuracy of \proposal and FedMatch according to helper selection parameters when $m=5$.}
    \label{fig:helper}
\end{figure}

\begin{figure}[t]
    \centering
        \subfigure[With varying number of helpers ($|H_r|$) 
        ]{
            \includegraphics[height=.34\linewidth]{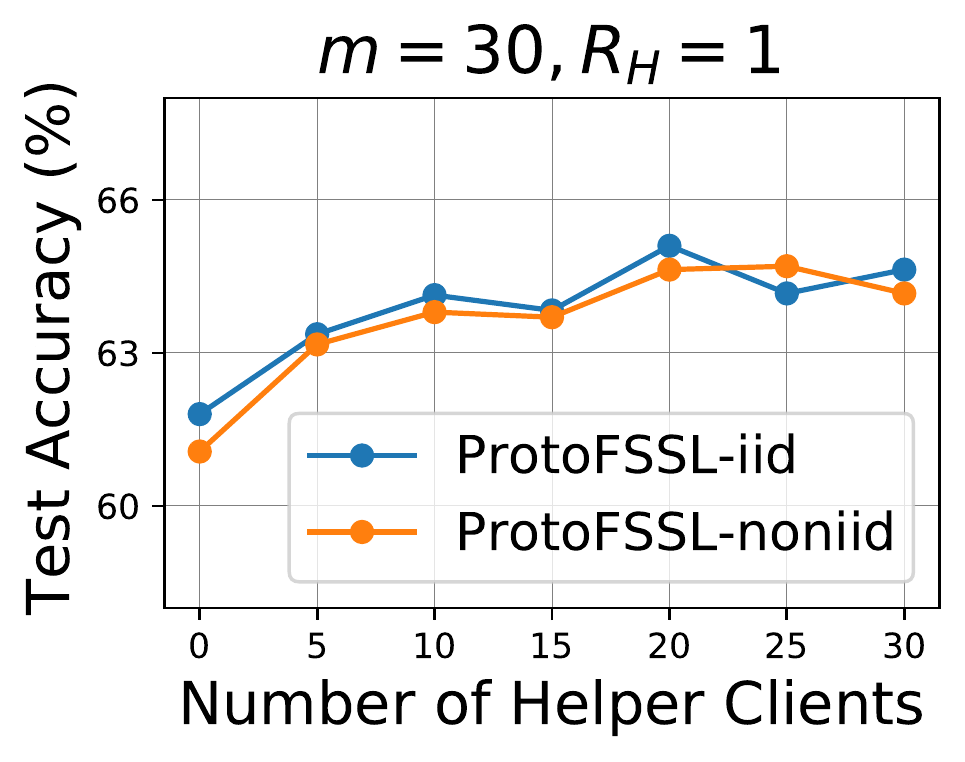}
        }
        \subfigure[With varying helper update interval ($R_H$) 
        ]{
            \includegraphics[height=.34\linewidth]{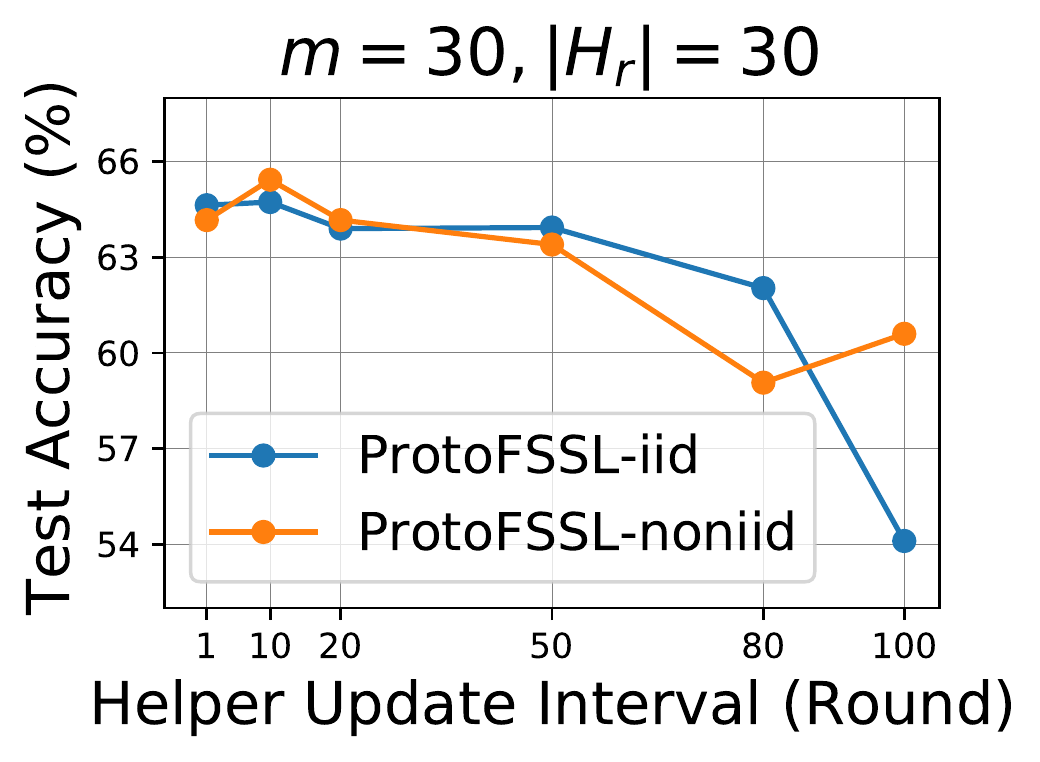}
        }
        \vspace{-2ex}
        \caption{Test Accuracy of \proposal on CIFAR-10 according to helper selection parameters when $m=30$.}
        \label{fig:helper_30}
\end{figure}

\textbf{Accuracy.}
\cref{table:accuracies} compares the test accuracy for \proposal and baselines. We set the FedMatch parameters to $|H_r|=2$ and $R_H=10$ following~\cite{jeong2021federated}. \hsk{Given that \proposal easily scales, we set its helper parameter to $|H_r|=5$. \proposal is also evaluated with various $|H_r|$ in Figures~\ref{fig:helper} and \ref{fig:helper_30}.}

We observe that \proposal significantly outperforms FedAvg and FedProx baselines, which do not utilize unlabeled data. This indicates that \proposal is able to extract knowledge from unlabeled data through pseudo-labeling with inter-client prototype exchanges. 
\proposal also outperforms both the simple combination of FL and SSL methods (i.e., FixMatch-FedAvg/FedProx) and FedMatch, demonstrating that the prototype-based knowledge sharing is superior to both no-sharing and model-based sharing. Notably, its performance degradation in the non-i.i.d settings is marginal, implying that inter-client prototypes effectively prevent clients from biased learning. In SVHN, \proposal is even comparable to the fully supervised learning cases.

In contrast, FedMatch results in the lowest accuracy, even lower than FedAvg and FedProx, which is also reported in \cite{lin2021semifed}. This shows the fragility of model-based knowledge sharing and disjoint learning. Note that in the original paper of FedMatch~\cite{jeong2021federated}, the reported test accuracy on CIFAR-10 is even worse than the lower-bound accuracy in \cref{table:accuracies}.

Lastly, to compare against FedRGD that does not provide inter-client knowledge sharing but adopts group normalization (GN) for mitigating local gradient diversity, we apply batch normalization (BN) to \proposal and observe that the performance of \proposal is improved and better than FedRGD on CIFAR-10. This shows that inter-client knowledge sharing and diversity mitigation techniques are complementary.

\begin{figure}[t]
    \centering
    \includegraphics[width=\linewidth]{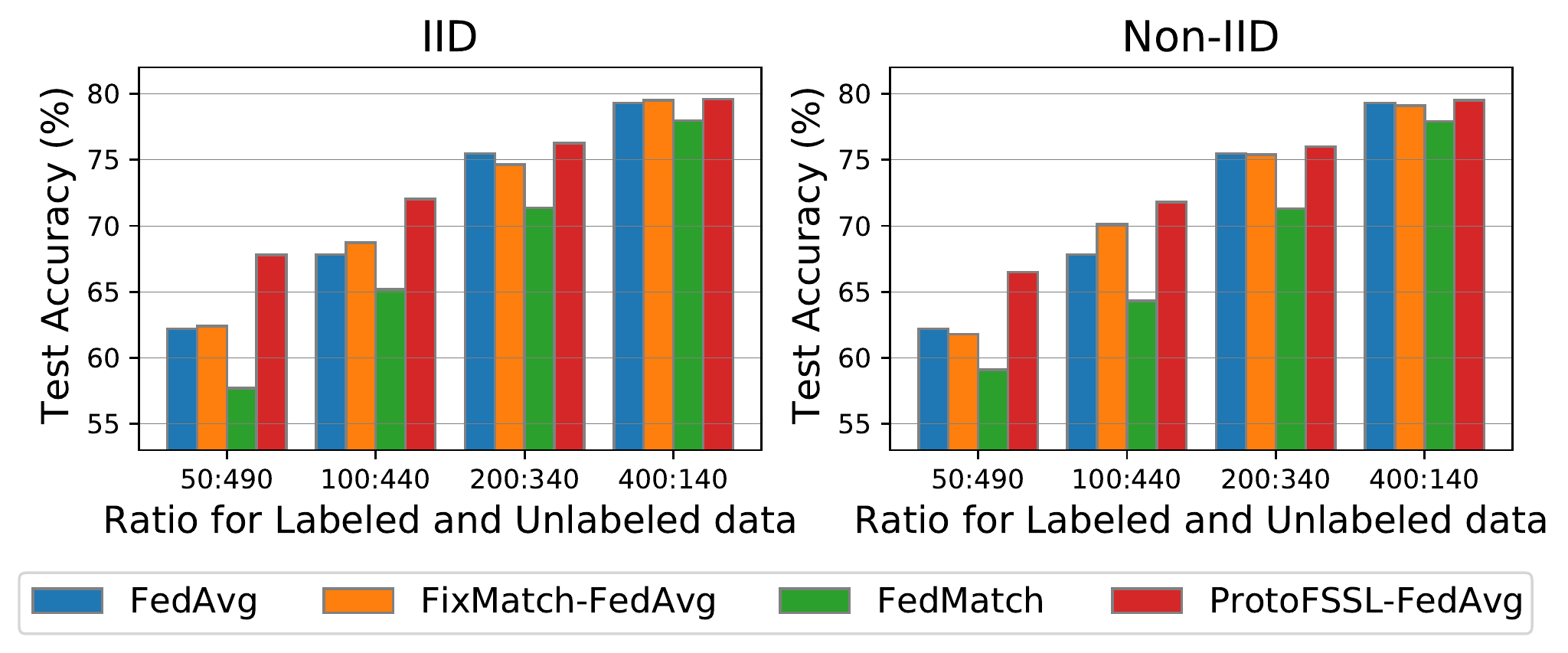}
    \vspace{-3ex}
    \caption{Test accuracy of various methods with different amount of labeled data in both i.i.d. and non-i.i.d. cases of CIFAR-10}
    \label{fig:label_ratio}
\end{figure}

\textbf{Impact of the Ratio of Labeled Data.}
In Figure~\ref{fig:label_ratio}, we plot the performance of various methods by adjusting the ratio of labeled vs. unlabeled data in the CIFAR-10 dataset. The results show that FedMatch has consistent lowest test accuracy, worse than the FedAvg baseline that does not utilize unlabeled data. This indicates that the way FedMatch utilizes unlabeled data can deteriorate training on labeled data.
In addition, FixMatch-FedAvg underperforms FedAvg in a few cases (e.g., the i.i.d. scenario with 200 labeled datapoints), meaning that simple combination of FL and SSL methods does not necessarily improve performance by using additional unlabeled data.
On the other hand, \proposal results in the best accuracy in all cases, showing the effectiveness of prototype-based knowledge sharing and pseudo-labeling. 

\textbf{Impact of Helper Client Selection.}
A main advantage of prototype-based knowledge sharing is in its low computation and communication costs, which enables \proposal to easily increase the number of helpers and share the prototypes frequently. To analyze the impact of the scalable design, we plot accuracy on CIFAR-10 relative to the number of helpers ($|H_r|$) and helper update interval ($R_H$) in Figures~\ref{fig:helper}(a) and \ref{fig:helper}(b). Note that although \proposal does not have a parameter for helper update interval as it updates the helpers in every round, in this experiment, we force the update interval to have different values for examining the effectiveness of prototypes.

The figures show that \proposal performance increases with a higher number of helpers and a lower helper update interval. The results verify that using more helpers \hsk{(i.e., less biased selection)} and providing the latest prototypes \hsk{(i.e., more accurate prototypes)} to clients contribute to performance improvement. Therefore, a lightweight design is important for improving accuracy without putting excessive burden on resource-constrained clients. 
On the other hand, FedMatch performance is almost unchanged by varying $|H_r|$ and $R_H$. Its model-based inter-client knowledge sharing and disjoint learning on unlabeled data is shown to provide marginal performance improvement.

To confirm the importance of scalability, we repeat the same experiments with more active clients ($m=30$) and plot \proposal performance in Figures~\ref{fig:helper_30}(a) and \ref{fig:helper_30}(b). The results show that to achieve the best performance, $|H_r|$ is desirable to be increased when having more active clients. This is because, as shown in~\cite{zhang2020improving}, more active clients increase gradient diversity among local models. Pseudo-labeling using knowledge of more various helper clients is helpful for addressing such gradient diversity. Again, the results prove the importance of a scalable design for inter-client knowledge sharing.

\vspace{-1ex}
\subsection{Ablation studies.}
\vspace{-1ex}
Next, we test how \proposal interacts with normalization methods and different base models. We include three normalization techniques; Batch Normalization (BN)~\cite{ioffe2015batch}, Group Normalization (GN) and Static Batch Normalization (sBN). 
While BN has been widely adopted, a number of studies showed that GN is superior to BN in FL settings with non-i.i.d data~\cite{wu2018group, reddi2021adaptive, hsieh2020non} and also in FSSL scenarios~\cite{zhang2020improving}. 
sBN is proposed in \cite{diao2021heterofl} as an alternative to BN, which reduces communication cost and improves privacy.

\cref{tab:ablations} shows the results. \proposal performance is improved with all three normalization methods, showing synergy between prototype-based knowledge sharing and normalization methods to improve model convergence.  
However, we observe that performance gain of using each normalization method for \proposal is different from that for FedAvg-SL. Although FedAvg provides the best accuracy with sBN and prior work has shown advantages of GN and sBN over BN, \proposal performs best when combined with traditional BN. This verifies that the effect of normalization methods on FL is hard to be generalized; performance gain would depend on how a FL framework is designed.

\begin{table}[t]
    \centering    
    \caption{Accuracy of \proposal combined with different techniques on CIFAR-10. We set $|M_r|=5$ and $R=300$.}
    \label{tab:ablations}
    \resizebox{.7\linewidth}{!}{%
    \begin{tabular}{c|cccc}
    \hline
    \multirow{2}{*}{Method}             & \multirow{2}{*}{Model}    & \multirow{2}{*}{Normalization} & \multicolumn{2}{c}{Test Acc. (\%)} \\
                                        &                           &                                & IID & Non-IID \\
    \hline
    \multirow{5}{*}{FedAvg-SL}   & ResNet9       & None          & 81.7   & 80.0 \\
                                        & ResNet9       & BN           & 87.8   & 86.8\\
                                        & ResNet9       & GN           & 85.7   & 82.0 \\
                                        & ResNet9       & sBN          & \textbf{88.8}   & \textbf{87.3} \\                                                                          \cline{2-5}
                                        & ResNet18      & sBN           & 88.8   & 87.5 \\
    \hline                          
    \multirow{5}{*}{\proposal}          & ResNet8       & None          & 66.1  & 65.6 \\
                                        & ResNet8       & BN            & \textbf{73.9}  & \textbf{73.4}\\
                                        & ResNet8       & GN            & 68.3  & 69.2\\
                                        & ResNet8       & sBN           & 73.6  & 68.9\\
                                        \cline{2-5}
                                        & ResNet17      & sBN           & 74.0  & 72.0  \\
    \hline
    \end{tabular}}
\end{table}

Lastly, \proposal is shown to work well for training a larger model ResNet17 without significant overfitting. Note that using ResNet18 can be overkill in CIFAR-10 given that FedAvg-SL does not show any performance improvement when using ResNet18 compared to ResNet9.

\vspace{-1ex}
\section{Discussion}
\label{sec:discussion}
\vspace{-1ex}
\subsection{Data Privacy}
\label{sec:privacy}
\vspace{-1ex}

While sharing prototypes is lightweight and thus preferred to sharing model weights, it could be vulnerable to the breach of data privacy as neural network activations could be inverted~\cite{mahendran2015understanding,dosovitskiy2016inverting,ulyanov2018deep} to reconstruct input data, either from a single or a few images~\cite{zhu2020deep}. 
While acknowledging the potential concerns, our framework mitigates the issue of data privacy as a class prototype is generated from an irreversible process~\cite{tan2021fedproto}: averaging the low-dimension (high-level) representations of samples from the class. \hsk{To improve privacy further, a client can perturb its prototypes when sharing (e.g., using Gaussian noise) and with the use of many helper clients, inter-client consistency regularization can still work robustly. A client having very few labeled samples for a class might also choose not to share its class prototype for privacy preservation, which does not harm model training since the prototype made out of very few samples is not reliable anyway. \proposal can also be integrated with various security techniques, such as differential privacy~\cite{mcmahan2017learning} and multi-party aggregation~\cite{bonawitz2017practical}.}  


\vspace{-1ex}
\subsection{Label Set Size}
\vspace{-1ex}

\hsk{While each prototype is a low-dimension vector, the number of prototypes increases with the number of classes, which might incur non-trivial communication cost in a dataset with large label set size. However, given that prototypes are usually much smaller than model size (13,000 times smaller in case of ResNet9 - a relatively lightweight CNN), applying the same network to CIFAR-100 (100 classes) or ImageNet (1000 classes), instead of CIFAR-10, results in not much difference. In addition, since a model naturally becomes larger to identify more classes accurately (e.g., more layers and more nodes in the last classification layer), prototype sharing can be still lighter than model sharing.} 
\vspace{-1ex}
\section{Conclusion}
\vspace{-1ex}

In this research, we study federated semi-supervised learning with heterogeneous data distribution, focusing on improving model convergence without increasing computation/communication burden on resource-constrained clients. We propose \proposal that imposes inter-client consistency regularization by exchanging lightweight prototypes between clients.
In experiments, we extensively evaluate \proposal on accuracy, computation/communication costs, the amount of  labeled data, helper and active client selection and relationship with normalization methods.
As the first attempt for applying prototypes for FSSL, we observe that the proposed \proposal is more effective than existing FSSL methods. The results reported in this paper provide insights for future exploration of prototype-based FSSL frameworks.

\printbibliography

\newpage
\appendix
\onecolumn

\section{Experimental Details}

\subsection{Training details}
We use RMSprop as an optimizer for training baseline models as well as \proposal. We also use L2 regularization on the weights of our base network. We fine-tune the proximal penalty parameter $\mu$ per each dataset when using FedProx. The hyperparameters used for training are summarized in \cref{tab:hyperparameters}.  $\eta$ is the base learning rate, $R$ is the number of total communication rounds, $\lambda_{L2}$ is L2 regularization factor, $M$ is the number of active clients per each round, $\mu$ is the proximal penalty parameter, $n_{S^L}$ is the number of samples per class for support set from labeled data, $n_{Q^L}$ is the number of samples per class for query set from labeled data, $n_{Q^U}$ is the number of samples for query set from unlabeled data, $H$ is the number of helpers, $\lambda_U$ is the parameter for unlabeled loss in \cref{eqn:loss} and $T$ is the \emph{temperature} used in sharpening, used in \cref{eqn:sharpening}. For non-IID distribution of unlabeled data, we follow the distribution used in FedMatch paper that controls the number of instances per class for each client. All of these hyperparameters are tuned using grid search.

For datasets, we use \textit{tensorflow.keras.datasets} and \textit{tensorflow\textunderscore datasets} libraries. However, as mentioned in \cref{sec:experiment}, we use different sizes of training, validation, and test sets. We follow splits used in \cite{jeong2021federated}, so we combined training set and test set in the library, shuffle them, and then divided according to the ratio we mentioned.

\begin{table}[h]
    \centering    
    \caption{Hyperparameters used in training for baselines and \proposal. More detailed hyperparameters could be found in our code.}
    \label{tab:hyperparameters}
    \resizebox{\linewidth}{!}{%
    \begin{tabular}{cccccccccccc}
    \toprule
    Methods         & $\eta$      & $R$     & $\lambda_{L2}$      & $M$     & $\mu$     & $n_{S^L}$ & $n_{Q^L}$ & $n_{Q^U}$ & $H$     & $\lambda_U$ & T \\
    \midrule
    FedAvg - SL         & 1e-3      & 300   & 1e-4              & 5     & -                 & -         & -         & -         & -     & -     & -    \\
    FedProx - SL        & 1e-3      & 300   & 1e-4              & 5     & \begin{tabular}{@{}c@{}}CIFAR-10: 1e-2 \\ SVHN: 1e-4 \\ STL-10: 1e-2 \end{tabular}    & -         & -         & -         & -     & -  & -    \\
    \proposal - FedAvg  & 1e-3      & 300   & 1e-4              & 5     & -                 & 1         & 2         & 100         & 5     & 0.3 & 0.5     \\
    \proposal - FedProx & 1e-3      & 300   & 1e-4              & 5     & \begin{tabular}{@{}c@{}}CIFAR-10: 1e-2 \\ SVHN: 1e-3 \\ STL-10: 1e-2 \end{tabular}    & 1         & 2         & 100         & 5     & 0.3  & 0.5     \\
    
    \bottomrule
    \end{tabular}}
\end{table}

\vspace{-2ex}
\subsection{Base network}
We used ResNet8 as our base network for \proposal, and ResNet9 for baseline models, following \cite{jeong2021federated}. Two models are identical except that ResNet8 does not have the last fully connected layer. Since our proposed method infers the probability for each class directly from prototypes, we do not need the last layer. This reduced network contributes to even smaller computation and communication overhead of \proposal. Skip connection is used in two places. (1) from the beginning of Conv3 to the end of Conv4, (2) from the beginning of Conv7 to the end of Conv8. The network architecture of ResNet8/ResNet9 is described in \cref{tab:netArchitecture}. 

\begin{table}[h]
    \centering
    \vspace{2ex}
    \caption{Architecture of base network - ResNet8 and ResNet9}
    \label{tab:netArchitecture}
    \begin{tabular}{cccccc}
    \toprule
    Layer   & Kernel size   & Input channel     & Output channel    & Stride    & Number of Parameters \\
    \midrule
    Conv1       & 3             & 3                 & 64                & 1         & 1728      \\
    Conv2       & 3             & 64                & 128               & 1         & 73,728     \\
    MaxPool1    & 2             & -                 & -                 & 2         & 0         \\
    Conv3       & 3             & 128               & 128               & 1         & 147,456    \\
    Conv4       & 3             & 128               & 128               & 1         & 147,456    \\
    Conv5       & 3             & 128               & 256               & 1         & 294,912    \\
    MaxPool2    & 2             & -                 & -                 & 2         & 0         \\
    Conv6       & 3             & 256               & 512               & 1         & 1,179,648   \\
    MaxPool3    & 2             & -                 & -                 & 2         & 0         \\
    Conv7       & 3             & 512               & 512               & 1         & 2,359,296   \\
    Conv8       & 3             & 512               & 512               & 1         & 2,359,296   \\
    MaxPool4    & 4             & -                 & -                 & 4         & 0         \\
    \hline
    \multicolumn{5}{c}{\textbf{ResNet8 total parameters}}                           & 6,563,520   \\
    \hline
    Softmax     & -             & 512               & 10                & -         & 5,120      \\
    \hline
    \multicolumn{5}{c}{\textbf{ResNet9 total parameters}}                           & 6,568,640   \\
    
    \bottomrule
    \end{tabular}
    \vspace{-3ex}
\end{table}

\vspace{-1ex}
\section{Further Analysis and Experimental Results}
\subsection{Alternative approach for prototypical training to use global prototype}
\begin{wrapfigure}{r}{0.6\textwidth}
    \begin{minipage}{0.6\textwidth}
        \vspace{-6ex}
        \begin{algorithm}[H]
           \caption{Alternative training approach using global prototypes}
           \label{alg:ProtoFSSL_alt}
        \begin{algorithmic}
            \STATE {\bfseries RunClient}($\theta^{r-1}, \{c_{j,k}\}_{k \in K, j \in H_r}$):
            \STATE Initialize $\theta^r_{i,0} \leftarrow \theta^{r-1}$
            \FOR{each local epoch $e = 1, 2, ..., E$}
            \FOR{each class $k \in K$}
                \STATE $ Q^L_{i,k} \leftarrow$ random sample from $D^L_{i,k}\setminus{S^L_{i,k}}$
                \STATE $ c_{i,k} \leftarrow \frac{1}{M_r}\sum_{j}{\{c_{j,k}\}_{j \in H_r}} $ 
            \ENDFOR
            \STATE Same as \cref{alg:ProtoFSSL}
            \ENDFOR
        \end{algorithmic}
        \end{algorithm}
        \vspace{-3ex}
    \end{minipage}
\end{wrapfigure}

In our proposed algorithm, we sampled both support set and query set from the labeled data, then the support set is used to make prototypes and the query set is used to train the network. But once we have prototypes from other clients, we can use them as a support set, similar to the method proposed in \cite{tan2021fedproto}. This allows labeled data owned by the local client to be fully leveraged as a query set. The client training part of the alternative algorithm is described in \cref{alg:ProtoFSSL_alt}. The test accuracy of this algorithm on CIFAR-10 dataset is 34.0\% and significantly lower than \proposal by 22.1\%p. This implies the support set plays an important role to extract useful knowledge from local labeled data.

\subsection{Impact of the number of samples from unlabeled data (\texorpdfstring{$n_Q^L$}{})}

\begin{wraptable}{r}{0.3\textwidth}
    \centering    
        \vspace{-6ex}
        \caption{Impact of varying $n_Q^L$ on CIFAR-10 dataset. We fix all other hyperparameters except $n_Q^L$ for the experiment.}
        \label{tab:numUnlabled}
        \resizebox{\linewidth}{!}{
        \begin{tabular}{ccc}
        \toprule
        \multirow{2}{*}{$n_Q^L$}       & \multicolumn{2}{c}{Test Acc.(\%)} \\
                                        & IID   & Non-IID                   \\
        \midrule
        10                              & 65.4  & 65.4  \\
        20                              & 65.5  & 65.7  \\
        30                              & 65.5  & 65.9  \\
        50                              & 66.7  & 66.7  \\
        100                             & 66.6  & 66.6  \\
        200                             & 66.7  & 66.5  \\
        
        \bottomrule
        \end{tabular}}
        \vspace{-5ex}
\end{wraptable}

In each local epoch on the client, query set($Q_i^U$) is sampled from the unlabeled dataset for training. We use $n_Q^L = 100$ for training. In this section, we analyze the effect of different $n_Q^L$s. Test accuracy on CIFAR-10 dataset is reported in \cref{tab:numUnlabled}. We fix all other parameters except $n_Q^L$ for the experiment. When the number of samples from unlabeled data is 10, the test accuracy is lower by 1.2\%p and 1.5\%p respectively, compared to the case of the number of samples from unlabeled data being 100. Test accuracy shows an increasing pattern as the number of samples from unlabeled data increases, and the increase is the largest when the number of samples increases from 30 to 50 in i.i.d case and 10 to 20 in non-i.i.d case. This result also supports that \proposal is effective in extracting knowledge from unlabeled data.

\subsection{Impact of local epoch (\texorpdfstring{$E$}{})}

We test the impact of different local epoch($E$) and the experimental result on CIFAR-10 i.i.d data is summarized in \cref{tab:localEpisode}. Increasing the local epoch from 1 to 10 improves the test accuracy by 15.9\%p, when there are 5 active clients and by 16.3\%p when there are 20 active clients. Interestingly, when the local epoch is only 1, test accuracy is higher when there are 20 active clients, than when there are 5 active clients by 8.6\%p - implying that even coarsely-learned models could improve the global model when there are more of them. 

    \begin{table}[H]
        \centering    
        \caption{Impact of local epochs $E$ on CIFAR-10 dataset, i.i.d case. We fix all other hyperparameters except $E$ and $M$ for the experiment.}
        \label{tab:localEpisode}
        \begin{tabular}{ccccc}
        \toprule
        \multicolumn{5}{c}{Test Acc.(\%)} \\
        \hline
        \multirow{2}{*}{Active clients($M$)}    & \multicolumn{4}{c}{$E$} \\
                                                & 1 & 2 & 5 & 10    \\
        \midrule
        5       & 30.2  & 41.4  & 63.9  & 66.1 \\
        20      & 38.8  & 42.7  & 64.7  & 65.1  \\
        
        \bottomrule
        \end{tabular}
    \end{table}

\subsection{Impact of adding noise to clients' prototypes}

Given that sharing prototypes might cause privacy leaks (\cref{sec:privacy}), we add noise to each prototype for improving privacy and analyze how the noisy prototypes impact convergence. 
Specifically, before sending their prototypes to the server, clients add noise from Gaussian distribution, $\mathcal{N}(0,\,\sigma^{2})$, to each value of the prototypes. For the experiments, we use \proposal with BN and same hyperparameters as used in \cref{table:accuracies}.

\begin{table*}[ht]
    \centering
    \caption{Test accuracy with nosie for different datasets. Method is \proposal (with BN) - FedAvg.}
    \label{table:noise_accuracy}
    \resizebox{\textwidth}{!}{%
    \begin{tabular}{c|ccc|ccc|cc}
    \hline
                       & \multicolumn{3}{c|}{CIFAR-10} & \multicolumn{3}{c|}{SVHN} & \multicolumn{2}{c}{STL-10}     \\
    \hline
    \multirow{2}{*}{std($\sigma$)}        & \multirow{2}{*}{L:U}      & \multicolumn{2}{c|}{Test Acc. (\%)} 
    & \multirow{2}{*}{L:U}      & \multicolumn{2}{c|}{Test Acc. (\%)} 
    & \multirow{2}{*}{L:U}      & \multirow{2}{*}{Test Acc. (\%)}        \\
    
                                &                           & IID & Non-IID
                                &                           & IID & Non-IID
                                &                           &                                       \\
    \hline
   -                     &                           & 73.9  & 73.4
                                &                           & 93.8  & 93.2
                                &                           & 78.2                                \\

    0.01                        & \multirow{5}{*}{50:490}   & 73.4  & 72.0
                                & \multirow{5}{*}{50:490}   & 90.8  & 90.7
                                & \multirow{5}{*}{100:980}  & 78.4                                  \\
    
   0.05                         &                           & 73.4  & 72.4
                                &                           & 90.9  & 90.6
                                &                           & 78.4                                  \\
    
    0.1                         &                           & 73.8  & 72.4
                                &                           & 91.1  & 91.2
                                &                           & 78.4                                  \\
    
    0.5                         &                           & 73.2  & 71.4
                                &                           & 90.9  & 91.2
                                &                           & 78.1                                 \\
                                
    5.0                         &                           & 74.1  & 73.6
                                &                           & 91.3  & 91.1
                                &                           & 77.6                                 \\

    10.0                        &                           & 47.5  & 45.9
                                &                           & 86.3  & 86.2
                                &                           & 62.6                                 \\                                
    \hline
    \end{tabular}%
    }
\end{table*}

\cref{table:noise_accuracy} shows the results, where the first row is the baseline \proposal without noise. The results show that when the standard deviation is not too large (i.e., 10.0), \proposal preserves model accuracy even with use of noisy prototypes. Our assumption for the robustness is that the original \proposal combines diverse prototypes from multiple clients for pseudo-labeling and the combined knowledge is likely to be less noisy than each individual prototype; noise can be canceled to some extent while combining multiple prototypes.




\end{document}